\documentclass{article}

\usepackage[nonatbib]{nips_2018}

\usepackage[utf8]{inputenc} 
\usepackage[T1]{fontenc}    
\usepackage{hyperref}       
\usepackage{url}            
\usepackage{booktabs}       
\usepackage{amsfonts}       
\usepackage{nicefrac}       
\usepackage{microtype}      
\usepackage{color}
\usepackage{graphicx}
\usepackage{caption}
\usepackage{subcaption}

\title{A Simple Method for
    Commonsense Reasoning}

\author{
  Trieu H. Trinh\thanks{Work done as a member of the Google Brain Residency program (\url{g.co/brainresidency}.)} \\
  Google Brain\\
  \texttt{thtrieu@google.com} \\
  \And Quoc V. Le \\
  Google Brain\\
  \texttt{qvl@google.com} \\
}

\begin{document}
\maketitle

\begin{abstract}
Commonsense reasoning is a long-standing challenge for deep learning. For example, it is difficult to use neural networks to tackle the Winograd Schema dataset~\cite{levesque2011winograd}. In this paper, we present a simple method for commonsense reasoning with neural networks, using unsupervised learning. Key to our method is the use of language models, trained on a massive amount of unlabled data, to score multiple choice questions posed by commonsense reasoning tests. On both Pronoun Disambiguation and Winograd Schema challenges, our models outperform previous state-of-the-art methods by a large margin, without using expensive annotated knowledge bases or hand-engineered features. We train an array of large RNN language models that operate at word or character level on LM-1-Billion, CommonCrawl, SQuAD, Gutenberg Books, and a customized corpus for this task and show that diversity of training data plays an important role in 
test performance. 
Further analysis also shows that our system successfully discovers important features of the context that decide the correct answer, indicating a good grasp of commonsense knowledge. 

\end{abstract}

\section{Introduction}





Although deep neural networks have achieved remarkable successes  (e.g.,~\cite{krizhevsky2012imagenet,taigman2014deepface,simonyan2014very,szegedy2015going,he2015delving,he2016deep,hinton2012deep,hannun2014deep,xiong2016achieving,chiu2017state,bahdanau2014neural,sutskever2014sequence,wu2016google,hassan2018achieving}),
their dependence on supervised learning has been challenged as  a significant weakness. This dependence prevents deep neural networks from being applied to problems where labeled data is scarce.
An example of such problems is common sense reasoning, such as the Winograd Schema Challenge~\cite{levesque2011winograd}, where the labeled set is typically very small, on the order of hundreds of examples. 
Below is an example question from this dataset:
\begin{itemize}
\item {\it The trophy doesn’t fit in the suitcase because \textbf{it} is too big. What is too big?\\
\textbf{Answer 0}: the trophy. \textbf{Answer 1}: the suitcase\\}
\end{itemize}
Although it is straightforward for us to choose the answer to be
{\it "the trophy"} according to our common sense, answering this type of question is a great challenge for machines
because there is no training data, or very little of it.

In this paper, we present a surprisingly simple method for common sense reasoning with Winograd schema multiple choice questions. Key to our method is th e use of language models (LMs), trained on a large amount of unlabeled data, to score multiple choice questions posed by the challenge and similar datasets. More concretely, in the above example, we will first substitute the pronoun (\emph{"it"}) with the candidates (\emph{"the trophy"} and \emph{"the suitcase"}), and then use LMs to compute the probability of the two resulting sentences ({\it "The trophy doesn’t fit in the suitcase because \textbf{the trophy} is too big."} and {\it "The trophy doesn’t fit in the suitcase because \textbf{the suitcase} is too big."}). The substitution that results in a more probable sentence will be the correct answer. 

\clearpage

On both Pronoun Disambiguation and Winograd Schema challenges, our method outperforms previous state-of-the-art methods by a large margin, without using expensive annotated knowledge bases or hand-engineered features. On a Pronoun Disambiguation dataset, PDP-60, our method achieves 70.0\% accuracy, which is better than the state-of-art accuracy of 66.7\%. On a Winograd Schema dataset, WSC-273, our method achieves 63.7\% accuracy, 11\% above that of the current state-of-art result (52.8\%)\footnote{Code to reproduce these results are available at \url{https://github.com/tensorflow/models/tree/master/research/lm_commonsense}.}

A unique feature of Winograd Schema questions is the presence of a special word that decides the correct reference choice. In the above example, {\it "big"} is this special word. When {\it "big"} is replaced by {\it "small"}, the correct answer switches to {\it "the suitcase"}. Although detecting this feature is not part of the challenge, further analysis shows that our system successfully discovers this special word to make its decisions in many cases, indicating a good grasp of commonsense knowledge.

\section{Related Work}
Unsupervised learning has been used to discover simple commonsense relationships. For example, Mikolov et al.~\cite{mikolov2013efficient,mikolov2013distributed} show that by learning to predict adjacent words in a sentence, word vectors can be used to answer analogy questions such as: Man:King::Woman:?. Our work uses a similar intuition that language modeling can naturally capture common sense knowledge. The difference is that Winograd Schema questions require more contextual information, hence our use of LMs instead of just word vectors. 

Neural LMs have also been applied successfully to improve downstream applications~\cite{dai2015semi,ramachandran2016unsupervised,peters2018deep,howard2018fine}. In~\cite{dai2015semi,ramachandran2016unsupervised,peters2018deep,howard2018fine}, researchers have shown that pre-trained LMs can be used as feature representations for a sentence, or a paragraph to improve NLP applications such as document classification, machine translation, question answering, etc. The combined evidence suggests that LMs trained on a massive amount of unlabeled data can capture many aspects of natural language and the world's knowledge, especially commonsense information. 

Previous attempts on solving the Winograd Schema Challenge usually involve heavy utilization of annotated knowledge bases, rule-based reasoning, or hand-crafted features~\cite{peng2015solving,bailey2015winograd,schuller2014tackling}. In particular, Rahman and Ng~\cite{rahman2012resolving} employ human annotators to build more supervised training data. Their model utilizes nearly 70K hand-crafted features, including querying data from Google Search API. Sharma et al.~\cite{sharma2015towards} rely on a semantic parser to understand the question, query texts through Google Search, and reason on the graph produced by the parser. Similarly, Sch{\"u}ller~\cite{schuller2014tackling} formalizes the knowledge-graph data structure and a reasoning process based on cognitive linguistics theories. Bailey et al.~\cite{bailey2015winograd} introduces a framework for reasoning, using expensive annotated knowledge bases as axioms.

The current best approach makes use of the skip-gram model to learn word representations~\cite{quanliu16winograd}. The model incorporates several knowledge bases to regularize its training process, resulting in Knowledge Enhanced Embeddings (KEE). A semantic similarity scorer and a deep neural network classifier are then combined on top of KEE to predict the answers. The final system, therefore, includes both supervised and unsupervised models, besides three different knowledge bases. In contrast, our unsupervised method is simpler while having significantly higher accuracy. Unsupervised training is done on text corpora which can be cheaply curated.

Using language models in reading comprehension tests also produced many great successes. Namely Chu et al.~\cite{ChuLM16} used bi-directional RNNs to predict the last word of a passage in the LAMBADA challenge. Similarly, LMs are also used to produce features for a classifier in the Store Close Test 2017, giving best accuracy against other methods~\cite{mostafazadeh2017lsdsem}. In a broader context, LMs are used to produce good word embeddings, significantly improved a wide variety of downstream tasks, including the general problem of question answering~\cite{peters2018deep,yu2018qanet}.

\section{Methods}

\begin{figure}[t!]
\centering\small
\includegraphics[width=0.95\columnwidth]{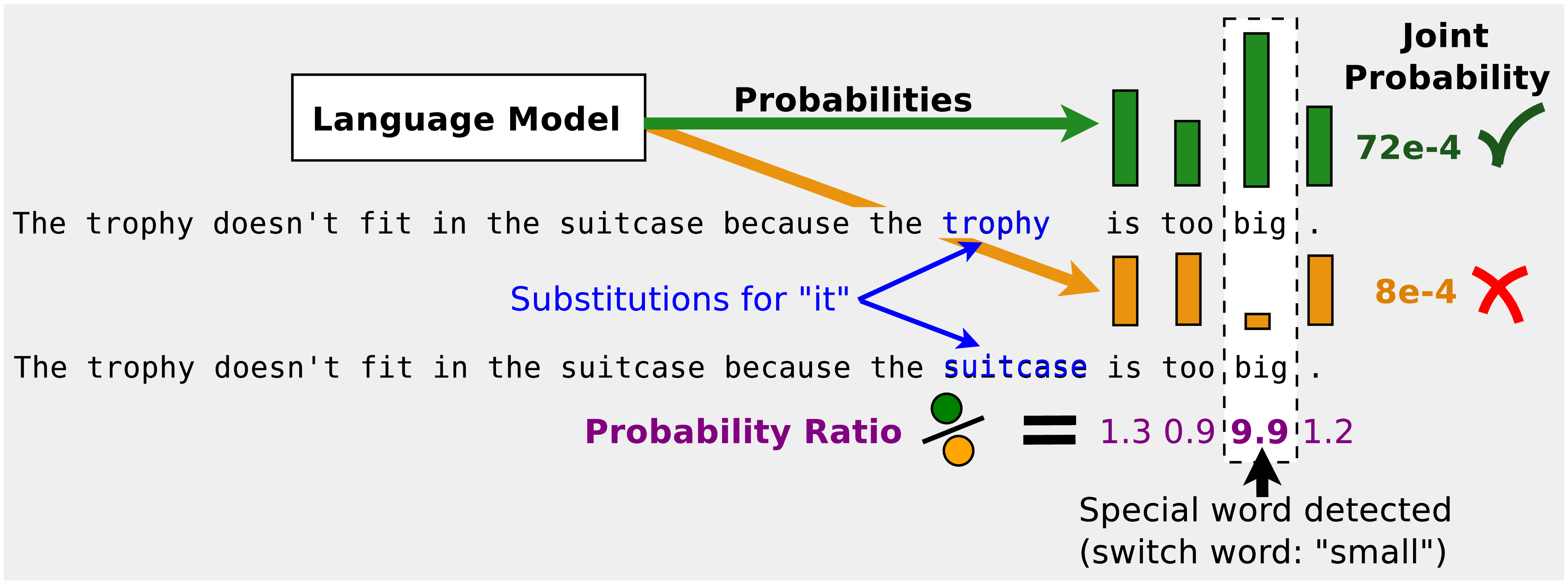}
\caption{Overview of our method and analysis. We consider the test {\it "The trophy doesn't fit in the suitcase because \textbf{it} is too big."} Our method first substitutes two candidate references {\it trophy} and {\it suitcase} into the pronoun position. We then use an LM to score the resulting two substitutions. By looking at probability ratio at every word position, we are able to detect {\it "big"} as the main contributor to {\it trophy} being the chosen answer. When {\it "big"} is switched to {\it "small"}, the answer changes to {\it suitcase}. This switching behaviour is an important feature characterizing the Winograd Schema Challenge.}
\label{fig:method}
\end{figure}

We first substitute the pronoun in the original sentence with each of the candidate choices. The problem of coreference resolution then reduces to identifying which substitution results in a more probable sentence. By reframing the problem this way, language modeling becomes a natural solution by its definition. Namely, LMs are trained on text corpora, which encodes human knowledge in the form of natural language. During inference, LMs are able to assign probability to any given text based on what they have learned from training data. An overview of our method is shown in Figure~\ref{fig:method}. 

Suppose the sentence $S$ of $n$ consecutive words has its pronoun to be resolved specified at the $k^{th}$ position: $S = \{w_1, .., w_{k-1}, w_{k} \equiv p, w_{k+1}, .., w_{n}\}$. We make use of a trained language model $P_\theta(w_t | w_{1}, w_2, .., w_{t-1})$, which defines the probability of word $w_t$ conditioned on the previous words $w_1, ..., w_{t-1}$.  
The substitution of a candidate reference $c$ in to the pronoun position $k$ results in a new sentence $S_{w_k\leftarrow c}$ (we use notation $w_k \leftarrow c$ to mean that word $w_k$ is substituted by candidate $c$). We consider two different ways of scoring the substitution:

\begin{itemize}
    \item $Score_{full}(w_k\leftarrow c) = P_{\theta}(w_1, w_2, ..., w_{k-1}, c, w_{k+1}, ..., w_n)$
\end{itemize}
which scores how probable the resulting full sentence is, and
\begin{itemize}
    \item $Score_{partial}(w_k \leftarrow c) = P_{\theta}(w_{k+1}, ..., w_n | w_1, ..., w_{k-1}, c) $
\end{itemize}
which scores how probable the part of the resulting sentence following $c$ is, given its antecedent. In other words, it only scores a part $S_{w_k\leftarrow c}$ conditioned on the rest of the substituted sentence. An example of these two scores is shown in Table~\ref{tab:scoring}. In our experiments, we find that \emph{partial} scoring strategy is generally better than the naive \emph{full} scoring strategy.

\begin{table}[h!]
  \scriptsize
  \caption{Example of \emph{full} and \emph{partial} scoring for the test {\it"The trophy doesn’t fit in the suitcase because \textbf{it} is too big."} with two reference choices {\it"the suitcase"} and {\it"the trophy"}.}
  \label{tab:scoring}
  \centering
  \begin{tabular}{l|l}
    \toprule
   $c$ = \textbf{the suitcase} &  $Score_{full}(w_k\leftarrow "the~suitcase")= P($The trophy doesn't fit in the suitcase because \textbf{the suitcase} is too big$)$ \\
    & $Score_{partial}(w_k\leftarrow "the~suitcase") = P($is too big$|$ The trophy doesn't fit in the suitcase because \textbf{the suitcase}$)$\\
    \midrule
    $c$ = \textbf{the trophy} & $Score_{full}(w_k\leftarrow "the~trophy") = P($The trophy doesn't fit in the suitcase because \textbf{the trophy} is too big$)$ \\& $Score_{partial}(w_k\leftarrow "the~trophy") = P($is too big$|$ The trophy doesn't fit in suitcase because \textbf{the trophy}$)$\\
    \bottomrule
  \end{tabular}
\end{table}


\section{Experimental settings}

In this section we describe tests for commonsense reasoning and the LMs used to solve these tasks. We also detail training text corpora used in our experiments.


\paragraph{Evaluation on Commonsense Reasoning Tests.}  We conduct experiments to evaluate our methods on two tasks: Pronoun Disambiguation Problems and Winograd Schema Challenge. These two tasks have been proposed as potential alternatives to the Turing Test, specifically targeting its potential weaknesses and inadequacy~\cite{levesque2011winograd}.

On the former task, we use the original set of 60 questions (PDP-60) as the main benchmark\footnote{\url{https://cs.nyu.edu/faculty/davise/papers/WinogradSchemas/PDPChallenge2016.xml}}. Later analysis augments this test with 62 questions from the development set to avoid bias presented in the original smaller set.\footnote{\url{http://commonsensereasoning.org/disambiguation.html}} The second task (WSC-273) is qualitatively much more difficult\footnote{\url{https://cs.nyu.edu/faculty/davise/papers/WinogradSchemas/WSCollection.xml}}. Its recent best reported result is only 3\% of accuracy above random guess~\cite{quanliu16winograd}. This task consists of 273 questions and is designed to work against techniques such as traditional linguistic restrictions, common heuristics or simple statistical test over text corpora ("\textit{Google-proof}")~\cite{levesque2011winograd}.

\paragraph{Recurrent language models.} We consider two types of recurrent LMs, one processes word inputs and the other processes character inputs. Their output layer, however, is constructed to only produce word outputs, allowing both types of input processing to join in ensembles. Namely, our LMs predict a distribution over a large vocabulary (800K words) at each time step, using a softmax layer. 
Following~\cite{rafal16lm}, we employ importance sampling at the softmax layer with 8,192 negative samples for each mini-batch to significantly speed up training. We use two layers of LSTM~\cite{hochreiter1997long} with 8,192 hidden units and a projection layer to a smaller dimensionality at output gates for faster processing.

For models that process words, we use a big embedding look up matrix with vocabulary size 800K and embedding size 1,024. For character-level input, we use a vocabulary size of 256 characters and embedding size 16. Characters in the same word are concatenated and used as input at a single time step. The resulting character embedding is processed using eight convolutions before going into the LSTM layers. More details about our LMs can be found in Appendix~\ref{sec:lms}.



\paragraph{Training text corpora.} We perform experiments on several different text copora to examine the effect of training data type on test accuracy. Namely, we consider LM-1-Billion, CommonCrawl\footnote{We evaluate all models trained on CommonCrawl after approximately 10-billion words are consumed.}, SQuAD and Gutenberg Books. For SQuAD, we collect context passages from the Stanford Question-Answering Dataset~\cite{rajpurkar2016squad} to form its training and validation sets accordingly. 

\section{Main results} \label{sec:exp}

Our experiments start with testing LMs trained on all text corpora with PDP-60 and WSC-273. Next, we show that it is possible to customize training data to obtain even better results.

\subsection{The first challenge in 2016: PDP-60}

We first examine unsupervised single-model resolvers on PDP-60 by training one character-level and one word-level LM on the Gutenberg corpus. 
In Table~\ref{tab:unsupervised}, these two resolvers outperform previous results by a large margin. For this task, we found \emph{full} scoring gives better results than \emph{partial} scoring. In Section~\ref{sec:better_partial}, we provide evidences that this is an atypical case due to the very small size of PDP-60.

\begin{table}[h!]
  \caption{Unsupervised single-model resolver performance on PDP-60}
  \label{tab:unsupervised}
  \centering\small
  \begin{tabular}{ll}
    \toprule
    Method & Accuracy \\
    \midrule
    Unsupervised Semantic Similarity Method (USSM) &   48.3 \% \\
    USSM + Cause-Effect Knowledge Base~\cite{quanliu16causecom} &  55.0 \% \\
    USSM + Cause-Effect + WordNet~\cite{miller1995wordnet} + ConceptNet~\cite{liu2004conceptnet} Knowledge Bases &  56.7 \% \\
    \midrule
    Char-LM-\emph{partial} & 45.0 \% \\
    Char-LM-\emph{full} & 53.3 \% \\
    Word-LM-\emph{partial} & {53.3 \%} \\
    \textbf{Word-LM-\emph{full}} & \textbf{60.0 \%} \\
    \bottomrule
  \end{tabular}
\end{table}

\begin{table}[h!]
  \caption{Unconstrained resolvers performance on PDP-60}
  \label{tab:full}
  \centering\small
  \begin{tabular}{ll}
    \toprule
    Method & Accuracy \\
    \midrule
    Patric Dhondt (WS Challenge 2016) & 45.0 \% \\
    Nicos Issak (WS Challenge 2016) & 48.3 \% \\
    Quan Liu (WS Challenge 2016 \textbf{- winner}) & 58.3 \% \\
    \midrule
    USSM + Supervised Deepnet &  53.3 \% \\
    USSM + Supervised Deepnet + 3 Knowledge Bases & 66.7 \% \\
    \midrule
    \textbf{Ensemble of 5 Unsupervised LMs-\emph{full}} & \textbf{70.0 \%} \\
    \bottomrule
  \end{tabular}
\end{table}

Next, we allow systems to take in necessary components to maximize their test performance. This includes making use of supervised training data that maps commonsense reasoning questions to their correct answer. Here we simply train another three variants of LMs on LM-1-Billion, CommonCrawl, and SQuAD and ensemble all of them. As reported in Table~\ref{tab:full}, this ensemble of five unsupervised models outperform the best system in the 2016 competition (58.3\%) by a large margin. Specifically, we achieve 70.0\% accuracy, better than the more recent reported results from Quan Liu et al (66.7\%)~\cite{quanliu16winograd}, who makes use of three knowledge bases and a supervised deep neural network.

\subsection{Winograd Schema Challenge}

On the harder task WSC-273, our single-model resolvers also outperform the current state-of-the-art by a large margin, as shown in Table~\ref{tab:winograd}. Namely, our word-level resolver achieves an accuracy of 56.4\%. By training another 4 LMs, each on one of the 4 text corpora LM-1-Billion, CommonCrawl, SQuAD, Gutenberg Books, and add to the previous ensemble, we are able to reach 61.5\%, nearly 10\% of accuracy above the previous best result. This is a drastic improvement considering this previous best system outperforms random guess by only 3\% in accuracy.

\begin{table}[h!]
  \caption{Accuracy on Winograd Schema Challenge}
  \label{tab:winograd}
  \centering\small
  \small
  \begin{tabular}{llc}
    \toprule
    Method & Accuracy\\
    \midrule
    Random guess & 50.0\% \\
    USSM + Knowledge Base & 52.0 \%\\
    USSM + Supervised DeepNet + Knowledge Base & 52.8 \%\\
    \midrule
    Char-LM-\emph{partial} & 51.3\%\\
    Char-LM-\emph{full} & 51.3\%\\
    Word-LM-\emph{partial} & 56.4\%\\
    Word-LM-\emph{full} & 53.8\%\\
    \textbf{Ensemble of 10 Unsupervised LMs-\emph{partial}} & \textbf{61.5 \%}\\
    \bottomrule
  \end{tabular}
\end{table}

This task is more difficult than PDP-60. First, the overall performance of all competing systems are much lower than that of PDP-60. Second, incorporating supervised learning and expensive annotated knowledge bases to USSM provides insignificant gain this time (+3\%), comparing to the large gain on PDP-60 (+19\%).\footnote{Our results so far have been with recurrent language models. As a comparison, we also trained a subword-level Transformer~\cite{vaswani2017attention} LM on Wikipedia texts and obtain competitive performance (58.3\% on PDP-60 and 54.1\% on WSC-273).}

\subsection{Customized training data for Winograd Schema Challenge}\label{sec:stories}

As previous systems collect relevant data from knowledge bases after observing questions during evaluation~\cite{rahman2012resolving,sharma2015towards}, we also explore using this option. Namely, we build a customized text corpus based on questions in commonsense reasoning tasks. It is important to note that this does not include the answers and therefore does not provide supervision to our resolvers.  
In particular, we aggregate documents from the CommonCrawl dataset that has the most overlapping n-grams with the questions. The score for each document is a weighted sum of $F_1(n)$ scores when counting overlapping n-grams:

$$Similarity\_Score_{document} = \frac{\sum_{n=1}^4nF_1(n)}{\sum_{n=1}^4n}$$

\begin{figure}[t!]
\centering
\includegraphics[width=1.0\columnwidth]{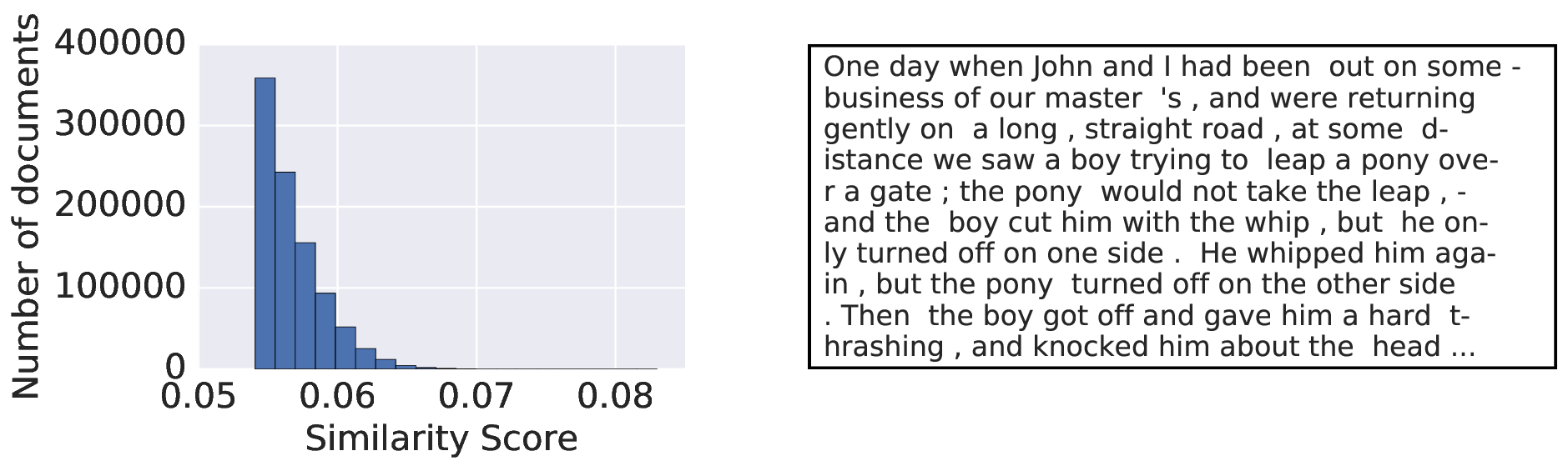}
\caption{\textbf{Left}: Histogram of similarity scores from top 0.1\% documents in CommonCrawl corpus, comparing to questions in Winograd Schema Challenge. \textbf{Right}: An excerpt from the document whose score is 0.083 (highest ranking). In comparison, a perfect score is of 1.0. Documents in this corpus contain long series of events with complex references from several pronouns.}
\label{fig:stories}
\end{figure}

The top 0.1\% of highest ranked documents is chosen as our new training corpus. Details of the ranking is shown in Figure~\ref{fig:stories}. This procedure resulted in nearly 1,000,000 documents, with the highest ranking document having a score of $8\times 10^{-2}$, still relatively small to a perfect score of $1.0$. We name this dataset STORIES since most of the constituent documents take the form of a story with long chain of coherent events.

We train four different LMs on STORIES and add them to the previous ensemble of 10 LMs, resulting in a gain of 2\% accuracy in the final system as shown in Table~\ref{tab:winograd_stories}. Remarkably, single models trained on this corpus are already extremely strong, with a word-level LM achieving 62.6\% accuracy, even better than the ensemble of 10 models previously trained on 4 other text corpora (61.5\%).

\begin{table}[h!]
  \caption{Accuracy on Winograd Schema Challenge, making use of STORIES corpus.}
  \label{tab:winograd_stories}
  \centering\small
  \small
  \begin{tabular}{llc}
    \toprule
    Method & Accuracy\\
    \midrule
    USSM + Supervised DeepNet + Knowledge Base & 52.8 \%\\
    \midrule
    Char-LM-\emph{partial} & 57.9\%\\
    Word-LM-\emph{partial} & 62.6\%\\
    \textbf{Ensemble of 14 LMs} & \textbf{63.7 \%}\\
    \bottomrule
  \end{tabular}
\end{table}

\section{Analysis}

\subsection{Discovery of special words in Winograd Schema}

We introduce a method to potentially detect keywords at which our proposed resolvers make decision between the two candidates $c_{correct}$ and $c_{incorrect}$. 
Namely, we look at the following ratio:

$$q_t = \frac{P_\theta(w_t | w_1, w_2, ..., w_{t-1}; w_k \leftarrow c_{correct})}{P_\theta(w_t | w_1, w_2, ..., w_{t-1}; w_k \leftarrow c_{incorrect})}$$

Where $1 \leq t \leq n$ for \emph{full} scoring, and $k +1 \leq t \leq n$ for \emph{partial} scoring. It follows that the choice between $c_{correct}$ or $c_{incorrect}$ is made by the value of $Q = \prod_tq_t$ being bigger than $1.0$ or not. By looking at the value of each individual $q_t$, it is possible to retrieve words with the largest values of $q_t$ and hence most responsible for the final value of $Q$.

\begin{figure}[h!]
\centering
\includegraphics[width=1.0\columnwidth]{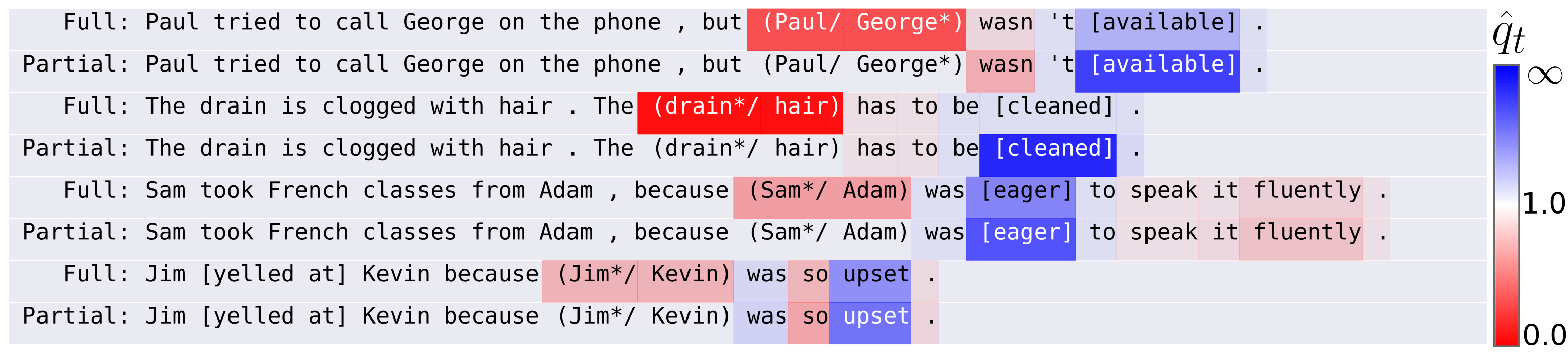}
\caption{A sample of questions from WSC-273 predicted incorrectly by \emph{full} scoring, but corrected by \emph{partial} scoring. Here we mark the correct prediction by an asterisk and display the normalized probability ratio $\hat{q}_t$ by coloring its corresponding word. It can be seen that the wrong predictions are made mainly due to $q_t$ at the pronoun position, where the LM has not observed the full sentence. \emph{Partial} scoring shifts the attention to later words and places highest $q$ values on the special keywords, marked by a squared bracket. These keywords characterizes the Winograd Schema Challenge, as they uniquely decide the correct answer. In the last question, since the special keyword appear before the pronoun, our resolver instead chose {\it "upset"}, as a reasonable switch word could be {\it "annoying"}.}
\label{fig:detect}
\end{figure}

We visualize the probability ratios $q_t$ to have more insights into the decisions of our resolvers. Figure~\ref{fig:detect} displays a sample of 
incorrect decisions made by \emph{full} scoring and is corrected by \emph{partial} scoring. 
Interestingly, we found $q_t$ with large values coincides with the special keyword of each Winograd Schema in several cases. Intuitively, this means the LMs assigned very low probability for the keyword after observing the wrong substitution. It follows that we can predict the keyword in each the Winograd Schema question by selecting top word positions with the highest value of $q_t$.

\begin{table}[h!]
  \caption{Accuracy of keyword detection from forward and backward scoring by retrieving top-2 words with the highest value of $q_t$}
  \label{tab:detect}
  \centering
  \small
  \begin{tabular}{lcc}
    \toprule
       & Resolution accuracy & Special word retrieved  \\
    \midrule
    Forward scoring & 63.7\% & 97 / 133 \\
    Backward scoring & 58.2\% & 18 / 45 \\
    \bottomrule
  \end{tabular}
\end{table}

For questions with keyword appearing before the reference, we detect them by backward-scoring models. Namely, we ensemble 6 LMs, each trained on one text corpora with word order reversed. This ensemble also outperforms the previous best system on WSC-273 with a remarkable accuracy of 58.2\%. Overall, we are able to discover a significant amount of special keywords (115 out of 178 correctly answered questions) as shown in Table~\ref{tab:detect}. This strongly indicates a correct understanding of the context and a good grasp of commonsense knowledge in the resolver's decision process.

\subsection{\emph{Partial} scoring is better than \emph{full} scoring.} \label{sec:better_partial}

In this set of experiments, we look at wrong predictions from a word-level LM. With \emph{full} scoring strategy, we observe that $q_t$ at the pronoun position is most responsible for a very large percentage of incorrect decisions as shown in Figfure~\ref{fig:detect} and Table~\ref{tab:parts}. For example, with the test {\it "The trophy cannot fit in the suitcase because \textbf{it} is too big."}, the system might return {\it $c_{incorrect} = $"suitcase"} simply because {\it $c_{correct} = $ "trophy"} is a very rare word in its training corpus and therefore, is assigned a very low probability, overpowering subsequent $q_t$ values. 

\begin{table}[h!]
  \caption{Error analysis from a single model resolver. Across all three tests, \emph{partial} scoring corrected a large portion of wrong predictions made by \emph{full} scoring. In particular, it corrects more than 62.7\% of wrong predictions on the Winograd Schema Challenge (WSC-273).}
  \label{tab:parts}
  \centering
  \small
  \begin{tabular}{lccc}
    \toprule
    \textbf{Data name} & \textbf{Wrong prediction} & \textbf{Corrected} & \textbf{Correction percentage} \\
    \midrule
    PDP-60 &  30 &  10  & 33.3\% \\
    PDP-122 &  55 &  33  & 60.0\% \\
    WSC-273 & 102  &  64 & 62.7\% \\
    \bottomrule
  \end{tabular}
\end{table}

\begin{figure}[h!]
\centering
\includegraphics[width=\columnwidth]{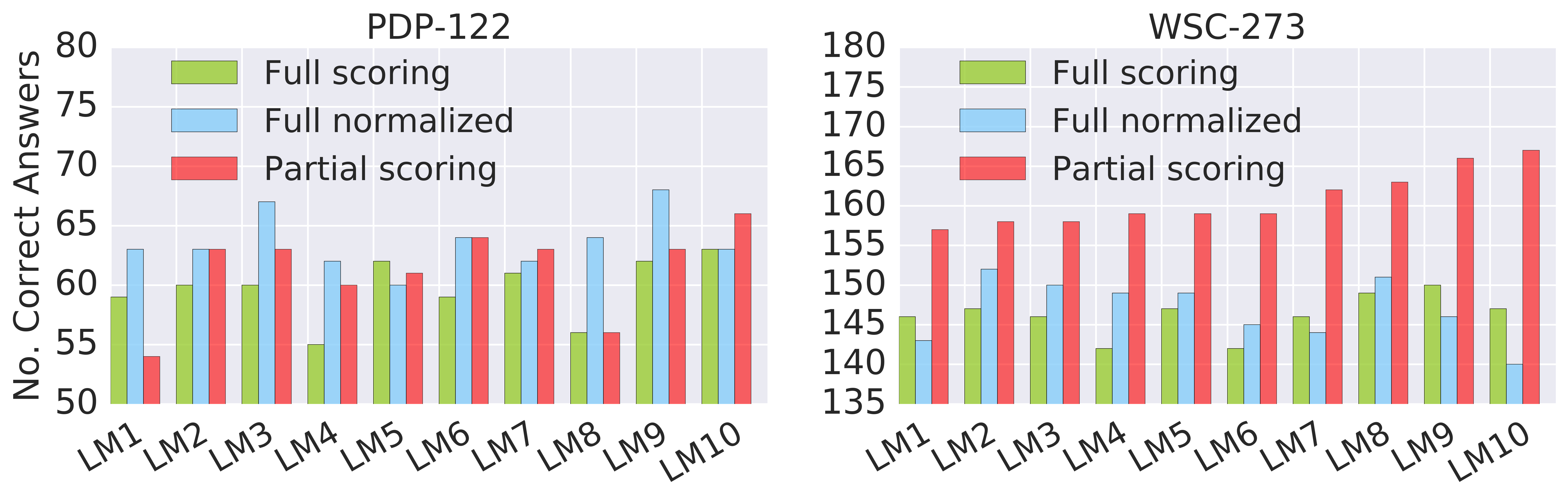}
\caption{Number of correct answers from 10 different LMs in three modes \emph{full}, \emph{full normalized} and \emph{partial} scoring. The second and third outperforms the first mode in almost all cases. The difference is most prominent on the largest test WSC-273, where \emph{partial} scoring outperforms the other methods by a large margin for all tested LMs.}
\label{fig:scoring}
\end{figure}

Following this reasoning, we apply a simple fix to \emph{full} scoring by normalizing its score with the unigram count of $c$: $Score_{full~normalized} = Score_{full} / Count(c)$. 
\emph{Partial} scoring, on the other hand, disregards $c$ altogether. As shown in Figure~\ref{fig:scoring}, this normalization fixes \emph{full} scoring in 9 out of 10 tested LMs on PDP-122. On WSC-273, the result is very decisive as \emph{partial} scoring strongly outperforms the other two scoring in all cases. Since PDP-122 is a larger superset of PDP-60, we attribute the different behaviour observed on PDP-60 as an atypical case due to its very small size.

\subsection{Importance of training corpus}

In this set of experiments, we examine the effect of training data on commonsense reasoning test performance. Namely, we train both word-level and character-level LMs on each of the five corpora: LM-1-Billion, CommonCrawl, SQuAD, Gutenberg Books, and STORIES. A held-out dataset from each text corpus is used for early stopping on the corresponding training data.

\begin{figure}[h!]
\centering
\includegraphics[width=1.0\textwidth]{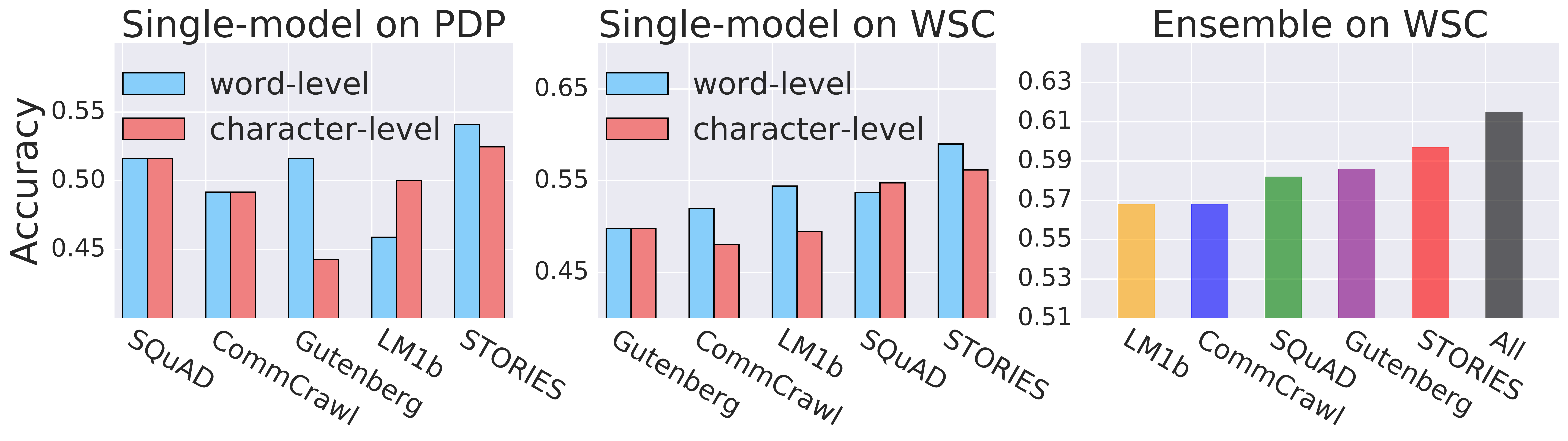}
\caption{\textbf{Left and middle}: Accuracy of word-level LM and char-level LM on PDP-122 and WSC-273 test sets, when trained on different text corpora. \textbf{Right}: Accuracy of ensembles of 10 models when trained on five single text corpora and all of them. A low-to-high ranking of these text corpora is LM-1-Billion, CommonCrawl, SQuAD, Gutenberg, STORIES.}
\label{fig:train_corpus}
\end{figure}

To speed up training on these large corpora, we first train the models on the LM-1-Billion text corpus. Each trained model is then divided into three groups of parameters: Embedding, Recurrent Cell, and Softmax. Each of the three is optionally transferred to train the same architectures on CommonCrawl, SQuAD and Gutenberg Books. The best transferring combination is chosen by cross-validation. 

Figure~\ref{fig:train_corpus}-left and middle show that STORIES always yield the highest accuracy for both types of input processing. We next rank the text corpora based on ensemble performance for more reliable results. Namely, we compare the previous ensemble of 10 models against the same set of models trained on each single text corpus. 
This time, the original ensemble trained on a diverse set of text corpora outperforms all other single-corpus ensembles including STORIES. 
This highlights the important role of diversity in training data for commonsense reasoning accuracy of the final system.

\section{Conclusion}

We introduce a simple unsupervised method for Commonsense Reasoning tasks. Key to our proposal are large language models, trained on a number of massive and diverse text corpora. The resulting systems outperform previous best systems on both Pronoun Disambiguation Problems and Winograd Schema Challenge. Remarkably on the later benchmark, we are able to achieve 63.7\% accuracy, comparing to 52.8\% accuracy of the previous state-of-the-art, who utilizes supervised learning and expensively annotated knowledge bases. We analyze our system's answers and observe that it discovers key features of the question that decides the correct answer, indicating good understanding of the context and commonsense knowledge. We also demonstrated that ensembles of models benefit the most when trained on a diverse set of text corpora. 

We anticipate that this simple technique will be a strong building block for future systems that utilize reasoning ability on commonsense knowledge.

\small
\bibliography{ref}

\begin{thebibliography}{10}

\bibitem{levesque2011winograd}
Hector~J Levesque, Ernest Davis, and Leora Morgenstern.
\newblock The winograd schema challenge.
\newblock In {\em AAAI spring symposium: Logical formalizations of commonsense
  reasoning}, 2011.

\bibitem{krizhevsky2012imagenet}
Alex Krizhevsky, Ilya Sutskever, and Geoffrey~E. Hinton.
\newblock Imagenet classification with deep convolutional neural networks.
\newblock In {\em Advances in Neural Information Processing Systems}, 2012.

\bibitem{taigman2014deepface}
Yaniv Taigman, Ming Yang, Marc'Aurelio Ranzato, and Lior Wolf.
\newblock Deepface: Closing the gap to human-level performance in face
  verification.
\newblock In {\em Proceedings of the IEEE conference on computer vision and
  pattern recognition}, pages 1701--1708, 2014.

\bibitem{simonyan2014very}
Karen Simonyan and Andrew Zisserman.
\newblock Very deep convolutional networks for large-scale image recognition.
\newblock {\em Advances in Neural Information Processing Systems}, 2015.

\bibitem{szegedy2015going}
Christian Szegedy, Wei Liu, Yangqing Jia, Pierre Sermanet, Scott Reed, Dragomir
  Anguelov, Dumitru Erhan, Vincent Vanhoucke, Andrew Rabinovich, et~al.
\newblock Going deeper with convolutions.
\newblock In {\em Proceedings of the IEEE Conference on Computer Vision and
  Pattern Recognition (CVPR)}, 2015.

\bibitem{he2015delving}
Kaiming He, Xiangyu Zhang, Shaoqing Ren, and Jian Sun.
\newblock Delving deep into rectifiers: Surpassing human-level performance on
  imagenet classification.
\newblock In {\em Proceedings of the IEEE international conference on computer
  vision}, pages 1026--1034, 2015.

\bibitem{he2016deep}
Kaiming He, Xiangyu Zhang, Shaoqing Ren, and Jian Sun.
\newblock Deep residual learning for image recognition.
\newblock In {\em Proceedings of the IEEE Conference on Computer Vision and
  Pattern Recognition (CVPR)}, pages 770--778, 2016.

\bibitem{hinton2012deep}
Geoffrey Hinton, Li~Deng, Dong Yu, George~E Dahl, Abdel-rahman Mohamed, Navdeep
  Jaitly, Andrew Senior, Vincent Vanhoucke, Patrick Nguyen, Tara~N Sainath,
  et~al.
\newblock Deep neural networks for acoustic modeling in speech recognition: The
  shared views of four research groups.
\newblock {\em IEEE Signal Processing Magazine}, 29(6):82--97, 2012.

\bibitem{hannun2014deep}
Awni Hannun, Carl Case, Jared Casper, Bryan Catanzaro, Greg Diamos, Erich
  Elsen, Ryan Prenger, Sanjeev Satheesh, Shubho Sengupta, Adam Coates, et~al.
\newblock Deep speech: Scaling up end-to-end speech recognition.
\newblock {\em arXiv preprint arXiv:1412.5567}, 2014.

\bibitem{xiong2016achieving}
Wayne Xiong, Jasha Droppo, Xuedong Huang, Frank Seide, Mike Seltzer, Andreas
  Stolcke, Dong Yu, and Geoffrey Zweig.
\newblock Achieving human parity in conversational speech recognition.
\newblock {\em arXiv preprint arXiv:1610.05256}, 2016.

\bibitem{chiu2017state}
Chung-Cheng Chiu, Tara~N Sainath, Yonghui Wu, Rohit Prabhavalkar, Patrick
  Nguyen, Zhifeng Chen, Anjuli Kannan, Ron~J Weiss, Kanishka Rao, Katya Gonina,
  et~al.
\newblock State-of-the-art speech recognition with sequence-to-sequence models.
\newblock {\em arXiv preprint arXiv:1712.01769}, 2017.

\bibitem{bahdanau2014neural}
Dzmitry Bahdanau, Kyunghyun Cho, and Yoshua Bengio.
\newblock Neural machine translation by jointly learning to align and
  translate.
\newblock In {\em International Conference on Learning Representations}, 2015.

\bibitem{sutskever2014sequence}
Ilya Sutskever, Oriol Vinyals, and Quoc~V Le.
\newblock Sequence to sequence learning with neural networks.
\newblock In {\em Advances in Neural Information Processing Systems}, pages
  3104--3112, 2014.

\bibitem{wu2016google}
Yonghui Wu, Mike Schuster, Zhifeng Chen, Quoc~V Le, Mohammad Norouzi, Wolfgang
  Macherey, Maxim Krikun, Yuan Cao, Qin Gao, Klaus Macherey, et~al.
\newblock Google's neural machine translation system: Bridging the gap between
  human and machine translation.
\newblock {\em arXiv preprint arXiv:1609.08144}, 2016.

\bibitem{hassan2018achieving}
Hany Hassan, Anthony Aue, Chang Chen, Vishal Chowdhary, Jonathan Clark,
  Christian Federmann, Xuedong Huang, Marcin Junczys-Dowmunt, William Lewis,
  Mu~Li, et~al.
\newblock Achieving human parity on automatic chinese to english news
  translation.
\newblock {\em arXiv preprint arXiv:1803.05567}, 2018.

\bibitem{mikolov2013efficient}
Tomas Mikolov, Kai Chen, Greg Corrado, and Jeffrey Dean.
\newblock Efficient estimation of word representations in vector space.
\newblock {\em arXiv preprint arXiv:1301.3781}, 2013.

\bibitem{mikolov2013distributed}
Tomas Mikolov, Ilya Sutskever, Kai Chen, Greg~S Corrado, and Jeff Dean.
\newblock Distributed representations of words and phrases and their
  compositionality.
\newblock In {\em Advances in neural information processing systems}, pages
  3111--3119, 2013.

\bibitem{dai2015semi}
Andrew~M Dai and Quoc~V Le.
\newblock Semi-supervised sequence learning.
\newblock In {\em Advances in Neural Information Processing Systems}, pages
  3079--3087, 2015.

\bibitem{ramachandran2016unsupervised}
Prajit Ramachandran, Peter~J Liu, and Quoc~V Le.
\newblock Unsupervised pretraining for sequence to sequence learning.
\newblock In {\em Conference on Empirical Methods in Natural Language
  Processing}, 2017.

\bibitem{peters2018deep}
Matthew~E Peters, Mark Neumann, Mohit Iyyer, Matt Gardner, Christopher Clark,
  Kenton Lee, and Luke Zettlemoyer.
\newblock Deep contextualized word representations.
\newblock In {\em Annual Conference of the North American Chapter of the
  Association for Computational Linguistics: Human Language Technologies},
  2018.

\bibitem{howard2018fine}
Jeremy Howard and Sebastian Ruder.
\newblock Fine-tuned language models for text classification.
\newblock {\em arXiv preprint arXiv:1801.06146}, 2018.

\bibitem{peng2015solving}
Haoruo Peng, Daniel Khashabi, and Dan Roth.
\newblock Solving hard coreference problems.
\newblock In {\em Proceedings of the 2015 Conference of the North American
  Chapter of the Association for Computational Linguistics: Human Language
  Technologies}, pages 809--819, 2015.

\bibitem{bailey2015winograd}
Dan Bailey, Amelia Harrison, Yuliya Lierler, Vladimir Lifschitz, and Julian
  Michael.
\newblock The winograd schema challenge and reasoning about correlation.
\newblock In {\em In Working Notes of the Symposium on Logical Formalizations
  of Commonsense Reasoning}, 2015.

\bibitem{schuller2014tackling}
Peter Sch{\"u}ller.
\newblock Tackling winograd schemas by formalizing relevance theory in
  knowledge graphs.
\newblock In {\em Fourteenth International Conference on the Principles of
  Knowledge Representation and Reasoning}, 2014.

\bibitem{rahman2012resolving}
Altaf Rahman and Vincent Ng.
\newblock Resolving complex cases of definite pronouns: the winograd schema
  challenge.
\newblock In {\em Proceedings of the 2012 Joint Conference on Empirical Methods
  in Natural Language Processing and Computational Natural Language Learning},
  pages 777--789. Association for Computational Linguistics, 2012.

\bibitem{sharma2015towards}
Arpit Sharma, Nguyen~Ha Vo, Somak Aditya, and Chitta Baral.
\newblock Towards addressing the winograd schema challenge-building and using a
  semantic parser and a knowledge hunting module.
\newblock In {\em IJCAI}, pages 1319--1325, 2015.

\bibitem{quanliu16winograd}
Quan Liu, Hui Jiang, Zhen{-}Hua Ling, Xiaodan Zhu, Si~Wei, and Yu~Hu.
\newblock Combing context and commonsense knowledge through neural networks for
  solving winograd schema problems.
\newblock {\em CoRR}, abs/1611.04146, 2016.

\bibitem{ChuLM16}
Zewei Chu, Hai Wang, Kevin Gimpel, and David~A. McAllester.
\newblock Broad context language modeling as reading comprehension.
\newblock {\em CoRR}, abs/1610.08431, 2016.

\bibitem{mostafazadeh2017lsdsem}
Nasrin Mostafazadeh, Michael Roth, Annie Louis, Nathanael Chambers, and James
  Allen.
\newblock Lsdsem 2017 shared task: The story cloze test.
\newblock In {\em Proceedings of the 2nd Workshop on Linking Models of Lexical,
  Sentential and Discourse-level Semantics}, pages 46--51, 2017.

\bibitem{yu2018qanet}
Adams~Wei Yu, David Dohan, Minh-Thang Luong, Rui Zhao, Kai Chen, Mohammad
  Norouzi, and Quoc~V Le.
\newblock Qanet: Combining local convolution with global self-attention for
  reading comprehension.
\newblock {\em arXiv preprint arXiv:1804.09541}, 2018.

\bibitem{rafal16lm}
Rafal J{\'{o}}zefowicz, Oriol Vinyals, Mike Schuster, Noam Shazeer, and Yonghui
  Wu.
\newblock Exploring the limits of language modeling.
\newblock {\em CoRR}, abs/1602.02410, 2016.

\bibitem{hochreiter1997long}
Sepp Hochreiter and J{\"u}rgen Schmidhuber.
\newblock Long short-term memory.
\newblock {\em Neural computation}, 9(8):1735--1780, 1997.

\bibitem{rajpurkar2016squad}
Pranav Rajpurkar, Jian Zhang, Konstantin Lopyrev, and Percy Liang.
\newblock Squad: 100,000+ questions for machine comprehension of text.
\newblock {\em arXiv preprint arXiv:1606.05250}, 2016.

\bibitem{quanliu16causecom}
Quan Liu, Hui Jiang, Zhen{-}Hua Ling, Si~Wei, and Yu~Hu.
\newblock Probabilistic reasoning via deep learning: Neural association models.
\newblock {\em CoRR}, abs/1603.07704, 2016.

\bibitem{miller1995wordnet}
George~A Miller.
\newblock Wordnet: a lexical database for english.
\newblock {\em Communications of the ACM}, 38(11):39--41, 1995.

\bibitem{liu2004conceptnet}
Hugo Liu and Push Singh.
\newblock Conceptnet—a practical commonsense reasoning tool-kit.
\newblock {\em BT technology journal}, 22(4):211--226, 2004.

\bibitem{vaswani2017attention}
Ashish Vaswani, Noam Shazeer, Niki Parmar, Jakob Uszkoreit, Llion Jones,
  Aidan~N Gomez, {\L}ukasz Kaiser, and Illia Polosukhin.
\newblock Attention is all you need.
\newblock In {\em Advances in Neural Information Processing Systems}, pages
  6000--6010, 2017.

\bibitem{duchi2011adaptive}
John Duchi, Elad Hazan, and Yoram Singer.
\newblock Adaptive subgradient methods for online learning and stochastic
  optimization.
\newblock {\em Journal of Machine Learning Research}, 12(Jul):2121--2159, 2011.

\end{thebibliography}
\bibliographystyle{unsrt}

\clearpage
\appendix

\section{Recurrent language models} \label{sec:lms}

The base model consists of two layers of Long-Short Term Memory (LSTM)~\cite{hochreiter1997long} with 8192 hidden units. The output gate of each LSTM uses peepholes and a projection layer to reduce its output dimensionality to 1024. We perform drop-out on LSTM's outputs with probability 0.25.

For word inputs, we use an embedding lookup of 800000 words, each with dimension 1024. For character inputs, we use an embedding lookup of 256 characters, each with dimension 16. We concatenate all characters in each word into a tensor of shape \emph{(word length, 16)} and add to its two ends the \emph{<begin of word>} and \emph{<end of word>} tokens. The resulting concatenation is zero-padded to produce a fixed size tensor of shape \emph{(50, 16)}. This tensor is then processed by eight different 1-D convolution (Conv) kernels of different sizes and number of output channels, listed in Table~\ref{tab:cnn}, each followed by a ReLU acitvation. 
The output of all CNNs are then concatenated and processed by two other fully-connected layers with highway connection that persist the input dimensionality. The resulting tensor is projected down to a 1024-feature vector. For both word input and character input, we perform dropout on the tensors that go into LSTM layers with probability 0.25.

\begin{table}[h!]
  \caption{One-dimensional convolutional layers used to process character inputs}
  \label{tab:cnn}
  \centering
  \small
  \begin{tabular}{lcccccccc}
    \toprule
      & Conv 1 & Conv 2 & Conv 3 & Conv 4 & Conv 5 & Conv 6 & Conv 7 & Conv 8 \\
    \midrule
    Kernel size & 1 & 2 & 3 & 4 & 5 & 6 & 7 & 7 \\
    Output channels & 32 & 32 & 64 & 128 & 256 & 512 & 1024 & 2048 \\
    \bottomrule
  \end{tabular}
\end{table}

We use a single fully-connected layer followed by a $Softmax$ operator to process the LSTM's output and produce a distribution over word vocabulary of size 800K. During training, LM loss is evaluated using importance sampling with negative sample size of 8192. This loss is minimized using the AdaGrad~\cite{duchi2011adaptive} algorithm with a learning rate of 0.2. All gradients on LSTM parameters and Character Embedding parameters are clipped by their global norm at 1.0. To avoid storing large matrices in memory, we shard them into 32 equal-sized smaller pieces. In our experiments, we used 8 different variants of this base model as listed in Table~\ref{tab:variants}.

\begin{table}[h!]
  \caption{All variants of recurrent LMs used in our experiments.}
  \label{tab:variants}
  \centering
  \small
  \begin{tabular}{ll}
    \toprule
    LM name  & Difference to base settings \\
    \midrule
    Word-LM 1 & Dropout rate 0.1 \\
    Word-LM 2 & Learning rate 0.05  \\
    Word-LM 3 & Residual connections around LSTM layers  \\
    Word-LM 4 & Project dimension 2048, embedding dimension 2048, One layer of LSTM \\
    Char-LM 1 & Embedding dimension 4096, project dimension 2048 \\
    Char-LM 2 & Embedding dimension 2048, project dimension 2048 \\
    Char-LM 3 & Embedding dimension 1024, learning rate 0.1, Residual instead of Highway connection\\
    Char-LM 4 & Learning rate 0.002, Embedding dimension 1024 \\
    \bottomrule
  \end{tabular}
\end{table}

In Table~\ref{tab:usage}, we listed all LMs and their training text corpora used in each of the experiments in Section~\ref{sec:exp}.

\begin{table}[h!]
  \caption{Details of LMs and their training corpus reported in our experiments.}
  \label{tab:usage}
  \centering
  \small
  \begin{tabular}{l|l}
    \toprule
    \textbf{Experiment}  & \textbf{LM variant / training corpus} \\
    \midrule
    Single models on PDP-60 & Word-LM 1/Gutenberg and Char-LM 1/Gutenberg \\
    \midrule
    Ensemble on PDP-60 & \textbf{Two single models on PDP-60 +}  Word-LM 2/SQuAD + \\
     & Char-LM 2/LM1B + Char-LM 3/CommonCrawl  \\
    \midrule
    Ensemble of 10 models & \textbf{Ensemble on PDP-60 +} \\ 
    on WSC-273    & Word-LM 1/Gutenberg {\it (different random seed)} + Word-LM 1/LM1B + \\
                  & Char-LM 4/Gutenberg + Char-LM 4/SQuAD + Char-LM 4/CommonCrawl \\
    \midrule
    Ensemble of 14 models & \textbf{Ensemble of 10 models on WSC-273 +}  \\
    on WSC-273 & Word-LM 1/STORIES + Char-LM 2/STORIES + \\
     & Word-LM 3/STORIES + Word-LM 4/STORIES \\
    \midrule
    Ensemble of 6 & Word-LM 1/Gutenberg + Word-LM 1/STORIES + \\
    backward-scoring models & Char-LM 4/CommonCrawl + Char-LM 4/SQuAD +\\
    on WSC-273 & Word-LM 4/LM1B + Char-LM 2/STORIES + \\
    \bottomrule
  \end{tabular}
\end{table}

\section{Data contamination in CommonCrawl}

Using the similarity scoring technique in section~\ref{sec:stories}, we observe a large amount of low quality training text on the lower end of the ranking. Namely, these are documents whose content are mostly unintelligible or unrecognized by our vocabulary. 
Training LMs for commonsense reasoning tasks on full CommonCrawl, therefore, might not be ideal. On the other hand, we detected and removed a portion of PDP-122 questions presented as an extremely high ranked document.

\end{document}